\documentclass[journal,10pt]{IEEEtran}

\usepackage{graphicx}
\usepackage{bbm}
\usepackage{tabularx}
\usepackage{subfig}
\usepackage{amsmath}
\usepackage{algorithmic}

\graphicspath{{./figs/}}

\usepackage[margin=10pt,labelfont={bf,rm},textfont={small,it}]{caption}

\begin{document}
\title{Relative Depth Order Estimation Using Multi-scale Densely Connected Convolutional Networks}

\author{Ruoxi Deng,
        Tianqi Zhao, 
        Chunhua Shen,
        Shengjun Liu\thanks{R. Deng and S. Liu are with Central South University, China.
        T. Zhao is with Tsinghua University, China.
        C. Shen is with the University of Adelaide, Australia. This work was done when R. Deng was visiting the University 
        of Adelaide.
         }
}

\markboth{Relative depth order estimation}%
{Deng \MakeLowercase{\textit{et al.}}:}

\maketitle
\begin{abstract}

We study the problem of estimating the relative depth order of point pairs in a monocular
image.
Recent advances \cite{Zoran_2015_ICCV,Zhou_2015_ICCV} mainly focus on using deep convolutional neural networks (DCNNs)
to learn and infer the ordinal information from multiple contextual information of the points pair
such as global scene context, local contextual information, and the locations.
However,  it remains unclear how much each context contributes to  the task.
To address this, we first examine the contribution of each context cue
\cite{Zoran_2015_ICCV, Zhou_2015_ICCV} to the performance in the context of  depth order estimation.
We find out the local context surrounding the points pair contributes the most and the global scene context helps little.
Based on the findings, we propose a simple method, using a multi-scale densely-connected network to tackle the task.
Instead of learning the global structure, we dedicate to explore the local structure by learning to regress from regions of multiple
sizes around the point pairs. Moreover, we use the recent densely connected network \cite{HuangG} to encourage substantial feature reuse
as well as deepen our network to boost the performance.
We show in experiments that the results of our approach is on par with or better than
the state-of-the-art methods with the benefit of using only a small number of training data.

\end{abstract}


\IEEEpeerreviewmaketitle

\section{Introduction}

\IEEEPARstart{T}{he} depth ordinal information of two objects (points) in an image is an important visual cue for many computer vision tasks such as objects classification \cite{Enzweiler,SmithP} and semantic segmentation \cite{ZhangZ,Hildreth,KowdleA}.
The objective is to know which one is closer or further (or at the same depth) to the camera,
given a pair of pixels.             %
To estimate relative depth order,  traditional methods mainly depends on objects' boundary and junction
characteristics such as T-junction, convexity/concavity and inclusion \cite{calderero2013recovering,visa2014precision,JiaZ}.
The accuracy of these methods is limited.
Recently, Convolutional Neural Networks (CNNs) have achieved remarkable success on many vision tasks such as object recognition
\cite{krizhevsky2012imagenet,Simonyan14c,He2015} and semantic segmentation \cite{LinG,ChenLC}.
	Motivated by the powerful visual representation
 	and  generalization capability,
 	recent works \cite{Zoran_2015_ICCV,Zhou_2015_ICCV} of depth estimation have also used
 	CNNs to estimate the ordinal information between the point pairs, and demonstrated superior performance.
	In \cite{Zoran_2015_ICCV,Zhou_2015_ICCV},
 both methods attempt to explore multiple features,
 which include  the appearance of the points, the local contextual information, the global scene context and so on.
 The idea is to use the visual cues as much as possible to improve the models' performance.
 Moreover, they both apply the multi-stream network structure.  %
 Zhou et al. \cite{Zhou_2015_ICCV}  concatenates all the convolutional features.
In contrast, Zoran et al.'s \cite{Zoran_2015_ICCV} network applies hierarchical
concatenation of the convolutional features---the global feature first concatenates
with the RoI mask and is fed into a fully connected layer,
then concatenates with the other convolutional features and the masks.

\begin{figure}[!t]
\centering
\includegraphics[width=\linewidth]{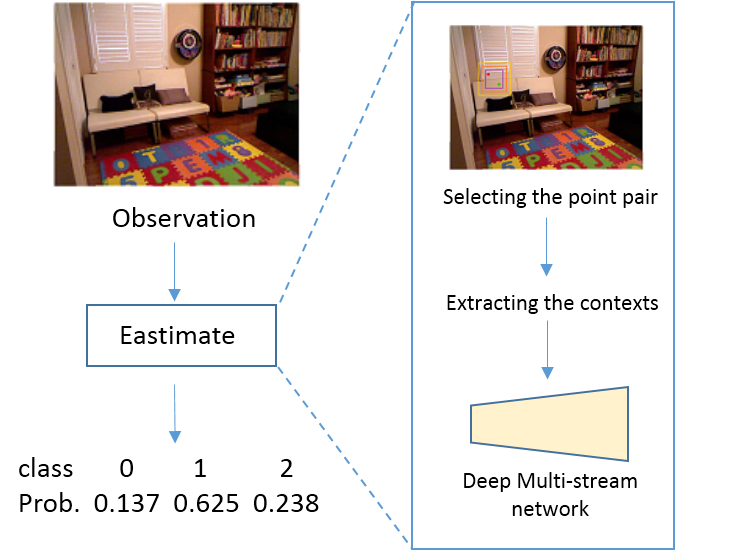}
\caption{The overall pipeline of estimating the depth order of a pair of points.
Given a pair of points, we extract its local contextual information and feed it to the proposed model
to perform the prediction.
The output of the model is the probability of a three-way classification,
which are three ordinal relationships ``at the same depth", ``further" and ``closer".}
\label{fig1}
\end{figure}

Thus, the studies of the recent works have mainly focused on combining various contextual information to train a network,
yet without demonstrating if each feature is useful.
In this work, we attempt to achieve two objectives:
1) empirically examine the contribution of each context cue;
2) and present a practical model to estimate the ordinal depth information.
As we show in the next sections,
such an exploration has resulted in several interesting findings.

\textbf{The global feature vs.\ the multi-scale local features}.
Following the insights presented in  \cite{Zoran_2015_ICCV, Zhou_2015_ICCV}, it makes sense to take advantage of
more types of the contextual information for improving the accuracy of the model.
However, neither of the them offers an analysis of the contributions of each cue.
It is crucial to find if each cue plays an important role in the model. For an ineffective cue, we can remove it to
 make the model simpler. Motivated by this, we conduct an experiment to examine the effectiveness of each cue. Our result shows that 
 the global scene context makes the least contribution in terms of the performance of the model. 
 We provide the detailed explanation in Section \ref{exame}.

\begin{figure}[!t]
\centering
\includegraphics[width=\linewidth]{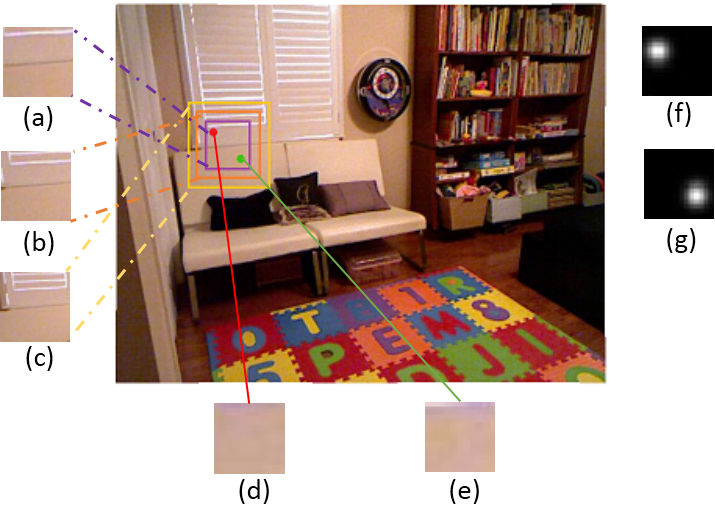}
\caption{ The different contextual information we use for training. (a)$\sim$(e) are the local contextual information, among which (a), (b) and (c) are the multi-scale features we proposed. (d) and (e) are the appearances of patch 1 and patch 2. (f) and (g) are the location information in the term of mask which represents their locations in the purple bounding box.  }
\label{multi-feat des}
\end{figure}

Consider two points (red and green) located in the purple bounding box in Fig.~\ref{multi-feat des}. 
If we remove the bookcase or the carpet from the scene, it would not affect the depth order of the points. We argue that the global structure of the scene is not necessarily useful for the task because the global information is redundant. 
Instead, the local context surrounding the points is critical.
 The local context contains abundant monocular cues such as occlusion, shadows, texture gradient and so on \cite{depth1,depth3, Fowlkes}, which determine the relative depth of the objects. 
 In this paper, we make CNN learn these relative depth cues by feeding the local background context surrounding the points. In particular, the context is in the form of multiple scales. Multi-scale features have an extended history application in computer vision and recently it has also been found very useful in the tasks like semantic segmentation \cite{LinG, ChenLC}, stereo vision \cite{chen2015deep} and high-quality natural images producing \cite{denton2015deep}, integrating with DCNN. Compared with the complex global structure, the local surrounding context is much simpler and easy to learn. Our experimental results show this simple change---that is, from learning the global context to learning the multi-scale local context---leads to a significant improvement in the performance.

\textbf{Deepening the model with DenseNet} Recently convolutional neural network has been witnessed to become deeper and deeper, from a few layers \cite{krizhevsky2012imagenet} to more than a thousand layers \cite{He2015}.  Very deep network structures such as VGG \cite{Simonyan14c}, highway network \cite{highway}, deep residual learning \cite{He2015} have demonstrated the superiority in many applications \cite{mao,Simonyan14c,ZagoruykoS,XieS}.

In this paper, we employ a novel deepening technique, namely densely connected network (DenseNet) \cite{HuangG}, to obtain powerful visual representation and improve the performance. The method has been reported to achieve 
 state-of-the-art performance on image classification tasks. Its principal characteristic is the dense connectivity. 
 That is, each layer of the structure is directly connected to every other layer in a feed-forward way. It takes advantage of the feature reuse and strengthens feature propagation. By using the densely connected network, we have $1.7\%$ percent accuracy improvement compared with our baseline model proposed in Section \ref{proposed} on the
  NYU Depth Dataset. We compare the DenseNet with the popular deepening technique, deep residual learning (ResNet) \cite{He2015} to demonstrate its advantages.

Last, we integrate the learned prior (the outputs of the proposed model which are the probabilities of three ordinal relationships) into the energy minimization proposed by Zoran et al.\ \cite{Zoran_2015_ICCV}, such that  
   we obtain the relative depth of the entire image from the ordinal estimates.
    The key difference is that we solve the minimization in the log space and  introduce a useful smoothness
     term, which improves the details of the recovered depth map.

In summary, our contributions are as follows.\\
\indent 1) We present a detailed experimental study   on the usefulness 
 the contextual information used in  prior works \cite{Zoran_2015_ICCV, Zhou_2015_ICCV}  by
   examining their effects for the task of depth order estimation.\\
\indent 2) We present a simple yet effective model, using the multi-scale framework and densely connected network, 
which makes the learning much easier and achieves  state-of-the-art performance. 
However, our method only uses hundreds of training images, while recent state-of-the-art methods \cite{ChenFYD16} usually 
used many more images (220K images) for training.\\
\indent 3) Last, we solve a constrained quadratic optimization problem similar to \cite{Zoran_2015_ICCV} to reconstruct the depth map from the ordinal estimations. 
We introduce a smoothness term to improve the result.

The rest of this paper is organized as follows. 
In Section \ref{rw} we summarize recent advances on ordinal relationship estimation and the CNN deepening techniques related to 
our work.  In Section \ref{main}, we  present  our method in detail, including the examination of the role of each context, describing the proposed model and how we reconstruct the depth map from the ordinal relationship of thousands of point pairs. 
In section \ref{experiment}, we provide quantities of experiments and analysis to validate the effectiveness of the proposed model.

\begin{figure*}[t]
\centering
\begin{minipage}{0.23\textwidth}
\subfloat[]{\includegraphics[width=\linewidth]{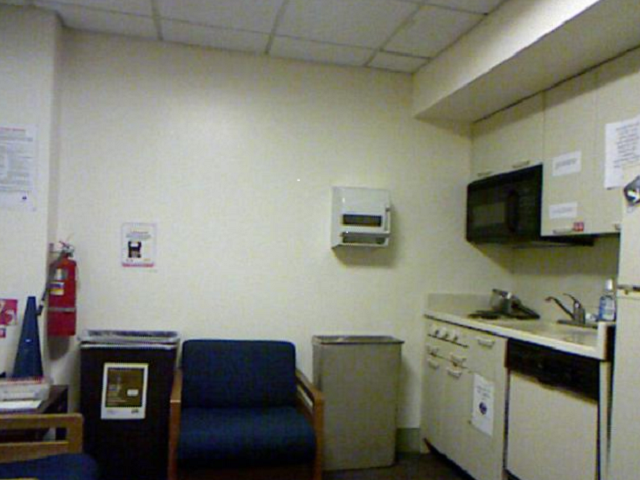}}
\end{minipage}
\begin{minipage}{0.23\textwidth}
\subfloat[]{\includegraphics[width=\linewidth]{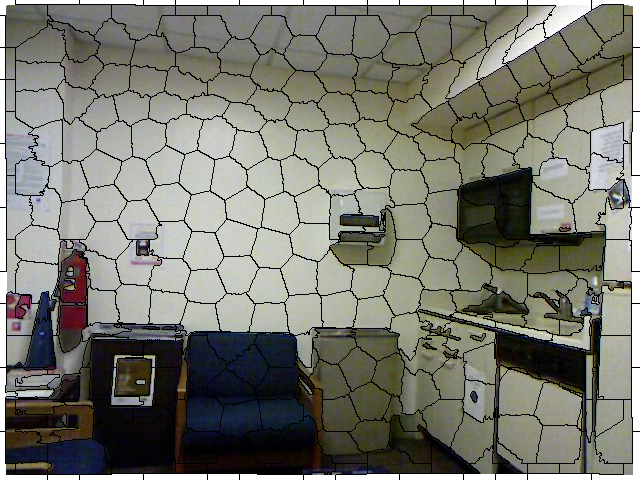}}
\end{minipage}
\begin{minipage}{0.23\textwidth}
\subfloat[]{\includegraphics[width=\linewidth]{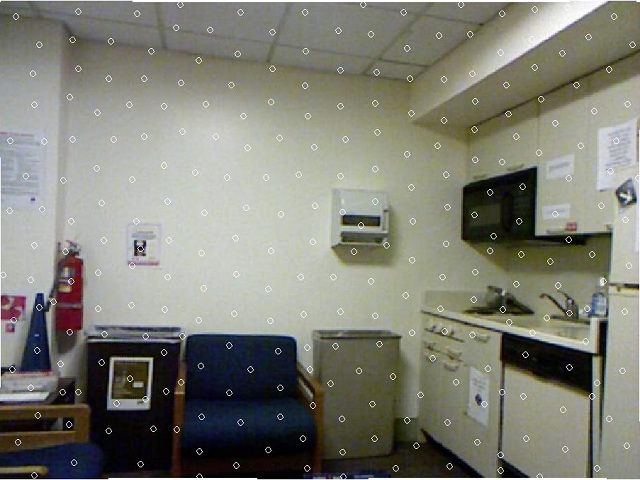}}
\end{minipage}
\begin{minipage}{0.23\textwidth}
\subfloat[]{\includegraphics[width=\linewidth]{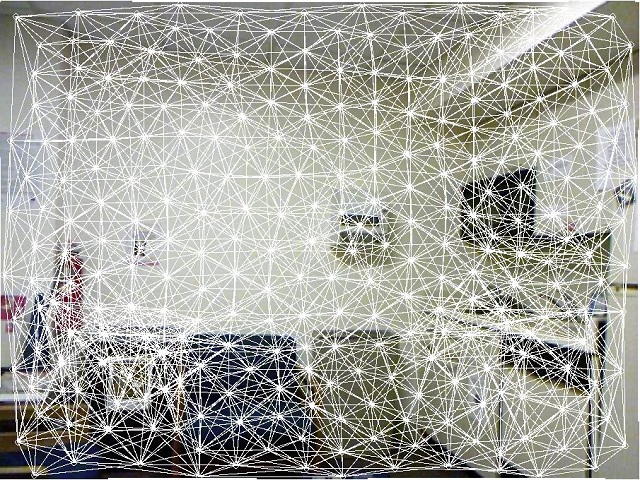}}
\end{minipage}
\caption{An example of how we extract and pair the points in a non-annotation depth dataset. 
(a) is RGB image from the NYU depth dataset; (b) The image is segmented into many superpixels. 
According to the superpixels, we find their centroids as the selected points and visualized in (c). 
We pair these points with their second-order neighbors in (d).}
\label{superpixel}
\end{figure*}

\section{Related works}
\label{rw}
In this section, we briefly review the works of monocular relative depth estimation, 
in particular, those using CNN. Furthermore, 
we  briefly review the characteristics of the current popular CNN deepening techniques.

\textbf{Depth ordinal relationship estimation in monocular images} 
Computer vision approaches handling monocular relative depth estimation were profoundly influenced by psycho-visual theory \cite{depth1,depth3}, which suggests T-junctions as one of the fundamental of monocular depth perception. Many works relied on developing the computational model to interpret and extract the T-junctions in an image \cite{calderero2013recovering,JiaZ, KowdleA,calderero2013recovering,visa2014precision}. 

Recently, due to the wide adoption of affordable  depth sensors, datasets such as the
 NYU Depth dataset \cite{silberman2012indoor} and KITTI dataset \cite{geiger2013vision,menze2015object} become available,
  thus leading to the trend of solving the problem as a supervised learning task using CNN. Pioneering work of Zoran et al.\ \cite{Zoran_2015_ICCV} proposed an end-to-end system to estimate the depth order of points pairs. Compared with another ordinal estimation work which also utilized CNN \cite{Zhou_2015_ICCV}, Zoran et al.\ add one more bounding box which contains the important visual cues for the task and a different location expression in their network structure. We consider such the design  very practical, which makes the visual cues  be automatically learned and inferred by CNN. Since the NYU and KITTI datasets have no direct annotations for the task, Chen et al.\ \cite{ChenFYD16} proposed a dataset named Depth in the Wild (DIW) which annotated the ordinal depth information between point pairs. Their deep model is an deep ranking model, and
    the output is a relative depth map.

Our network is an end-to-end system. The output is the same as in \cite{Zoran_2015_ICCV,Zhou_2015_ICCV}, which are three relationships, namely, ``further", ``closer" and ``the same". 
However, we employ different input context cues and a much deeper architecture. 
We describe the details in the next section.

\textbf{CNN deepening techniques} The number of the layers of CNN has dramatically increased in 
 recent years, from AlexNet \cite{krizhevsky2012imagenet}, VGG \cite{Simonyan14c}, Inception \cite{inception} to ResNet \cite{He2015}. The deepening of CNN is not merely repeating the `Convolution-Relu-Pooling' process. 
 With increasing the depth of a plain network, the performance of a deep model was often observed worse than a shallower model, due to the gradient vanishing and the optimization becoming underfit \cite{He2015}. 
 To address the issues, a well-designed structure is needed. In VGG \cite{Simonyan14c}, $3\times3$ filters are used throughout, and the entire network is divided into several blocks. Convolutional layers are stacked in the block. Between the blocks, the max pooling operation is used to reduce the feature map size. This  design highly influences  network structures proposed in
  recent years, as nowadays most of the structures use $3\times3$ convolutional filters  and the block by block structure.
  
   The characteristic of Inception network \cite{inception} is that their structure is not only deep but also very wide. In their block, several streams with different filter sizes ($1 \times 1$, $3 \times 3$) are applied. ResNet \cite{He2015} is the most successful network structure since it well alleviates the gradient vanishing problem.  It is worth mentioning 
   the highway network \cite{highway} because its underlying principle is similar to ResNet. 
   ResNet applies skip connections and sum up the feature map every two or three layers to enhance the information flow, which is termed residual learning. Relying on residual learning, ResNet can extend its depth to more than a thousand layers and still achieves impressive performance.

The densely connected network is a new deep structure proposed by Huang et al.\ \cite{HuangG}.
 It takes advantage of the above useful designs such as small filters ($3\times3$, $1\times1$), block by block structure (yet apply average pooling) and skip connection as well. In particular, its primary characteristic is the dense connectivity, which leads to the heavy feature reuse, information flow propagation, and good regularization. Considering these advantages, we apply the DenseNet in our proposed model. To verify our choice, we compare its effectiveness  against ResNet 
  and the baseline model, respectively. We  observe the improvement over the proposed baseline model. The details are shown in Section \ref{experiment}.

\begin{figure*}[!t]
\begin{minipage}{.23\linewidth}
\centering
\subfloat[]{\label{main:a}\includegraphics[scale=.45]{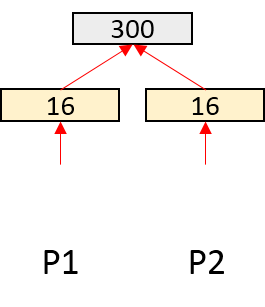}}

\end{minipage}%
\begin{minipage}{.23\linewidth}
\subfloat[]{\label{main:b}\includegraphics[scale=.45]{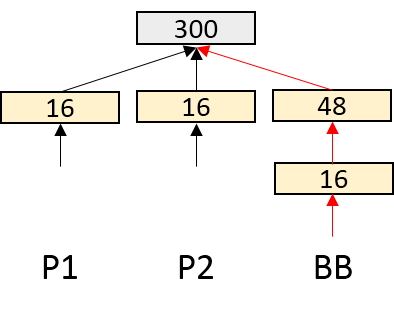}}
\end{minipage}
\begin{minipage}{.25\linewidth}

\subfloat[]{\label{main:c}\includegraphics[scale=.45]{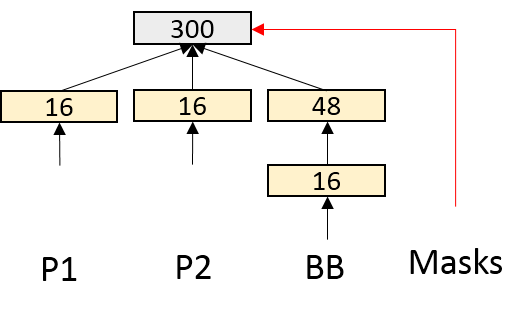}}
\end{minipage}
\begin{minipage}{.25\linewidth}

\subfloat[]{\label{main:d}\includegraphics[scale=.45]{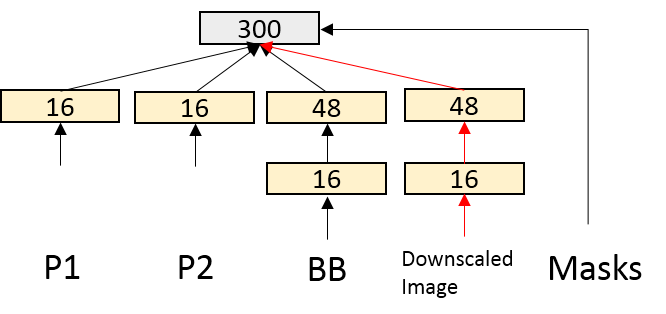}}
\end{minipage}
\caption{The models for the ablation experiments. We start from (a), which only keeps the streams of the patches. 
The red line denotes the newly added component in each model. 
The yellow, gray rectangle denotes the convolutional layer and the fully-connected layer, respectively. 
The number on it denotes the  dimension of the outputs. 
Details of the training settings are given in the Section \ref{experiment}. }
\label{fig:main}
\end{figure*}

\section{Estimation of the depth order of a pair of  points}
\label{main}
Estimating the depth order of point pairs is to explore the points' 3D relationship. 
It is very challenging, as the only information that we use is the 2D appearances. 
We propose to tackle this complex problem in multiple stages.  

Firstly, since the quality of input features plays an important role in the training process, we analyze the effectiveness of the input feature (context) used in the state-of-the-art works \cite{Zoran_2015_ICCV,Zhou_2015_ICCV} by conducting a series of ablation experiments. We observe several interesting findings in the experiments. Secondly, based on the findings, we present the multi-scale model to make better use of the local contextual information surrounding the points. The method is simple, which extracts three gradually increased bounding boxes around the points (shown in Fig.~\ref{multi-feat des}) and feeds the contents of the bounding boxes into a deep CNN, instead of the global scene context used in \cite{Zoran_2015_ICCV,Zhou_2015_ICCV}. 
Thirdly, motivated by the recent success of very deep CNN, we manage to obtain the better performance by using the DenseNet \cite{HuangG} to deepen the proposed model.  To provide a reference, we compare the performance of the DenseNet with the ResNet \cite{He2015} when they are both used to deepen the proposed model. Last, we reconstruct the depth map from the outputs of the proposed model to deliver an intuitive impression of the quality of the estimates.

Most of depth datasets such as the NYU dataset V2 \cite{silberman2012indoor} and KITTI \cite{geiger2013vision} do not
 provide the annotations for ordinal estimation. Thus before estimating the ordinal relationship, we need to determine which pairs of points 
 should  be sampled in an image, then extract the required contextual information according to the selected points. We employ the strategy proposed by Zoran et al.\ \cite{Zoran_2015_ICCV} to achieve the goal. We over segment an image into many superpixels \cite{achanta2012slic} and pick the centroid of the superpixels as the selected points. We then paired the points with their second-order neighbors to compare the depth order. The method is simple yet effective since a superpixel varies very smoothly and the centroid of the superpixel can represent the other points within it. Fig.~\ref{superpixel} depicts an example of the selected points in an image and how the points are paired. After extracting the context, we start to examine the effectiveness of the context.

\subsection{Examining the contextual information used for depth order estimation}
\label{exame}

The contextual information used in recent works \cite{Zoran_2015_ICCV,Zhou_2015_ICCV} has three types: the local contextual information, the global scene context, and the location information. As illustrated in Fig.~2, 
the local contexts are two patches on the comparison points and a bounding box surrounding the patches. The global context is a downscaled version of the input image. The location information that we use here is the mask. Note that in Zoran's model \cite{Zoran_2015_ICCV},
the bounding box is considered as the global scene context. We would like to label it  local context since it contains the local structure in an image.

\begin{figure*}[t]
\centering
\includegraphics[width=\linewidth]{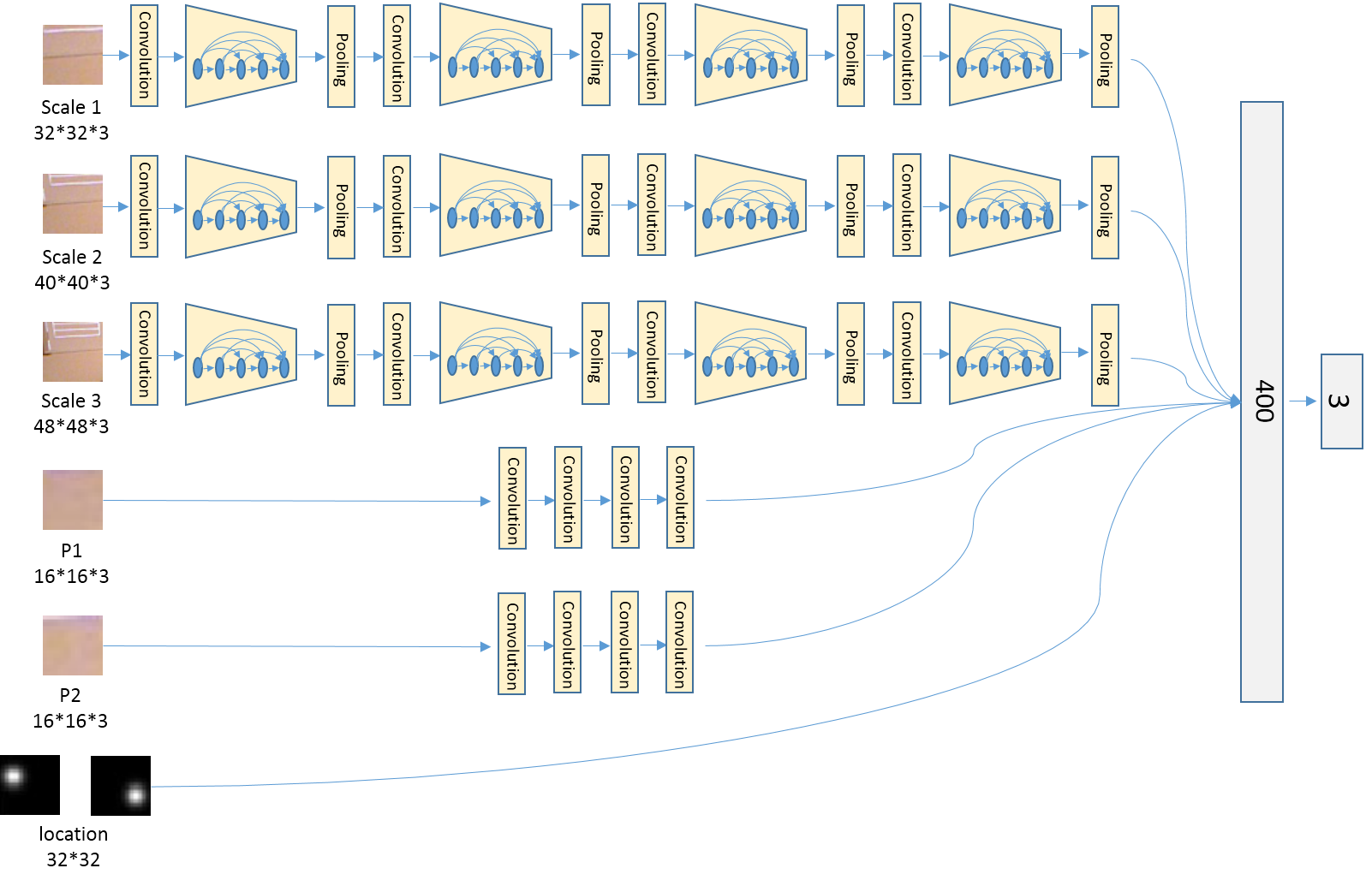}
\caption{The proposed deepened network structure. The trapezoid graphic denotes a DenseNet block inside. 
In the streams of multi-scale contexts, the convolutional operations are all padded. DenseNet uses average pooling between the adjacent blocks to downsample the feature maps in the streams. For the streams of the patch 1 and patch 2, we use strided convolutions with no pooling. If we remove the trapezoid graphics and average pooling layers in the multi-scale streams, the structure becomes the proposed baseline model before deepening in  Section \ref{proposed}. }
\label{network structure}
\end{figure*}

We carry out a series of ablation experiments to examine the effectiveness of these contexts for the task. 
Before starting the experiments, we need to complete several preparations. Firstly, we simplify Zoran's model to make it more intuitive. The simplified model that we utilize is shown in Fig.~3(d). It removes three fully connected layers from Zoran's model (the first one is next to the concatenation of ROI and Downscaled Image, the second one is next to BB, the last one is next to the fully connected layer which has 300 outputs). However, the number of convolutional layers and the parameter setting of the corresponding convolutional layer in the simplified model, such as the filter size and stride, are the same with Zoran's. 
We split the simplified model into several basic components according to their types. There basic components are the patches, the bounding box, the masks and the downscale version of the image.

We start the experiments from only one component and test its performance, then gradually add  other components one by one into the model and repeat the test. 
In this procedure, if the performance is improved, we can conclude the feature is likely to be effective for the task.
 If not, then the feature is not useful. Our test is conducted on the NYU depth dataset. Since the task is to learn the relationship between the points' appearances and the depth order, the first basic component that we keep in the model is the streams of two points' appearances (the patches). In the next  round of test,  we add the remaining components following the order of adding the bounding box---adding the masks---adding the downscaled image. In every round we test 10 times and select the most accurate result as the model's best performance. Everytime we test a new component, we fine tune the model from the previous model of the best performance. We present the results in the Table \ref{Ablation Experiments}.

\begin{table}[t!]
\renewcommand{\arraystretch}{1.2}
\caption{Ablation Experiments.}
\label{Ablation Experiments}
\centering
\newcommand{\specialcell}[2][c]{%
  \begin{tabular}[#1]{@{}c@{}}#2\end{tabular}}
\begin{tabular}{c||c}
\hline
Method & Accuracy of the predictions\\
\hline
Baseline (The patches) & 47.5\%\\

$+$Bounding Box  & 52.0\%\\

$+$Masks & 58.7\%\\
\specialcell{ $+$Downscaled Image \\ (simplified from Zoran el al.)  }& 58.1\%\\
Zoran et al. \cite{Zoran_2015_ICCV} & 59.6\%\\
\hline
\end{tabular}
\end{table}

We test the accuracy of the predictions to measure the performance of the models. We take the result of the basic component model as the baseline. We also test the accuracy of Zoran's model to show the difference in the performance between the original model and the simplified model. Our baseline (Fig. \ref{fig:main}(a)) achieves $47.5\%$. Adding the bounding box component (Fig. \ref{fig:main}(b)) leads to a significant improvement---the accuracy increases to $52.0\%$. Applying the masks (Fig. \ref{fig:main}(c)) gains a significant improvement as well:
 the accuracy further increases to $58.7\%$. However, the accuracy decreases to $58.1\%$ after adding the downscaled image (Fig.~\ref{fig:main}(d)). Zoran's model achieves $59.6\%$ accuracy which is almost $1\%$ more than our simplified model.

We see three findings from the experiments: 

1) The local background context and the location information are highly useful for the task; From the experiments, they contributes the most for the increase of the performance; 

2) The global structure may  not be required for the task, since adding the global scene context (the downscaled image), the performance shows
 a slight degradation; 

 3) The complex network structure, such as the hierarchy concatenation of the different streams \cite{Zoran_2015_ICCV} in the original Zoran's model, is also helpful for improving the performance.

Why would adding the global structure result in performance degradation? In theory, the global context has the global structure information which certainly contains the structure of the local context. Thus it should be helpful to boost the performance as well. 
We argue that the global structure may not be well
 learned, as its  structure, which only has two layers, may not be able to efficiently learn the semantic content of the scene. 

 A recent work \cite{Gonzalez-Garcia} by Gonzalez et al.\ investigates the responses of convolutional filters with semantic parts to analyze the internal representation of CNN. They argue that when CNNs handle the tasks like scene classification, it needs to be very deep, because the task is less related to object parts.  According to their finding, we can deepen the network to make more convolutional filters response to the semantic parts of the scene. To demonstrate this point of view, we deepen the stream of the downscaled image from two layers to four layers. We set the filter size of the four layers as $3\times3$ . The first two layers' stride is 2 and the last two layers' stride is 1. As we expected, the accuracy increases to $59.1\%$, which is slightly better than non-downscaled-image model in Fig. \ref{fig:main}(c). Gonzalez et al.\ suggest very deep architectures such as VGG \cite{Simonyan14c} for various tasks since it has abundant semantic parts for finetuning. Coincidentally, Chakrabarti et al.\ \cite{Chakrabarti} also utilize VGG to extract the coarse global structure of an image for absolute depth estimation. 
 As a comparison, training from scratch with such a shallow depth is not ideal.  %

\subsection{The proposed baseline model}
\label{proposed}
As discussed, extracting the global structure from an image is very challenging, and most of the structure is not helpful to estimate the depth order of the points. Thus we choose a different route. We first look at the promising results from the {\em deep visual correspondence embedding model} \cite{chen2015deep} for inspiration to make better use of the local context. The model from \cite{chen2015deep} achieves improved 
 accuracy of stereo matching by proposing an ensemble model of two patch scales. The large patches with richer information are less ambiguous, and the small patches have merits in details. The model combines the best of two worlds. We hypothesized that this may work better to learn the local structure from observing the local backgrounds at multiple scales with different sizes of contexts.

\begin{figure}[t!]
\begin{minipage}{.5\linewidth}
\centering
\subfloat[]{\label{main:a}\includegraphics[scale=.6]{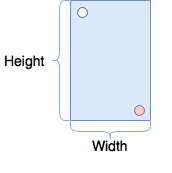}}
\end{minipage}%
\begin{minipage}{.5\linewidth}
\centering
\subfloat[]{\label{main:b}\includegraphics[scale=.4]{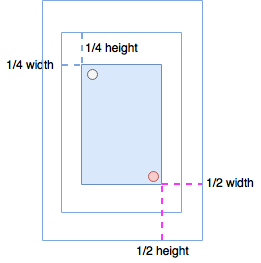}}
\end{minipage}
\caption{Illustration of three-scale bounding boxes used in the proposed baseline model. We first extract a rectangle surrounding the two points as the scale 1 bounding box which is shown in the left figure. We then produce the scale 2 and 3 bounding boxes via extending by 
 1/4 and 1/2 length of the width and height of the scale 1 
  as shown in the right figure.  }
\label{boundingboxes}
\end{figure}

Therefore, we propose a multi-scale model to incorporate multi-scale local background contexts. We add two larger bounding box scales with the fields of view of 2.25$\times$, 4$\times$ of the original bounding box (shown in Figure \ref{boundingboxes}) all surrounding the points pair. In each stream of the model, we utilize strided convolution of four layers for feature map downsampling. The output features from the patches and multi-scale local surroundings are concatenated with the location information and fed into a fully-connected layer with 400 outputs. We train the network end-to-end with the log-softmax loss.

The proposed model encourages to aggregate the multi-scale contextual information. We consider that three scales may be ideal for learning the crucial visual cues for the task. 
We set the model as a baseline in Section \ref{experiment}. In the next subsection, we show how to push the limit of the performance by deepening the proposed model. 

\subsection{Deepening with densely connected networks}
\label{deepened}
\begin{figure}
\begin{minipage}{.5\linewidth}
\centering
\subfloat[]{\label{main:a}\includegraphics[scale=.6]{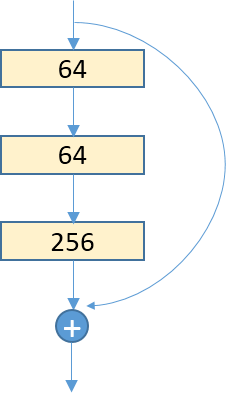}}
\end{minipage}%
\begin{minipage}{.5\linewidth}
\centering
\subfloat[]{\label{main:b}\includegraphics[scale=.6]{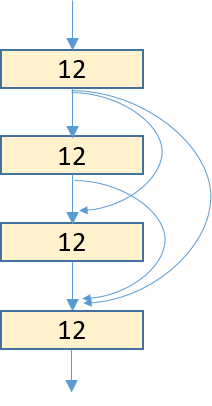}}
\end{minipage}
\caption{The structure of a ResNet block (left) and a DenseNet block (right). }
\label{ResDense}
\end{figure}

We employ the densely connected networks proposed by Huang et al.\ \cite{HuangG} to deepen the proposed model. 
As we have introduced, it is a very deep CNN structure. The particular design of the network is the dense connectivity. In a densely connected network, the dense connectivity denotes that for each layer, it connects with all the other layers in the network.
 See  Fig. \ref{ResDense}(b). The characteristic differs from the pattern of traditional CNN, of which each layer is only connected with its adjacent layers. In a densely connected network, the feature maps of the early layers are concatenated as the input of the later layer. For the $k_{th}$ layer, it receives the feature maps of all preceding layers, i.e.,
\[
 x_{k} = H_{k}([x_{0},x_{1},...,x_{k-1}])
\]
Here $H_{k}(.)$ denotes a non-linear transformation includes Convolution, Batch Normalization \cite{Szegedy} and rectified linear unit 
\cite{xavier}.
 As the number of the connections grows quadratically with the depth, Huang et al.\ provide two solutions to control the growth. The first one is that they present a hyper-parameter $k$ as the growth rate of the network. This growth rate $k$ is representing the number of the outputs in a layer. For instance, $k = 12$ denotes that all the layers in the densely connected network have 12 output feature maps. The other one is dividing the entire network into several blocks. Thus the dense connectivity is only applied within the blocks. Between the blocks, average pooling is used to downsample the feature map.

Now we show how to integrate the densely connected network in the proposed baseline model. 
As shown in Fig.~\ref{network structure}, the idea is simple: the model contains  four densely connected blocks with equal numbers of layers and average pooling layers  in the streams of the multi-scale local context inputs.  
In these deepened streams, all the convolutional layers use filters with kernel size $3 \times 3$, stride 1 with zero-padding to keep the feature map size fixed. The length of the layers in each densely connected block is 5 and the growth rate $k$ is 12. 
Note that, we have not deepened the streams of two points' appearance.  
Since the selected points for comparison often locate in the smooth area, the appearances of the points rarely have complex textures and edges. In contrast, the multi-scale local contexts have abundant visual cues and textures, which should apply the deep structure.

We demonstrate its benefit of enhanced 
 feature reuse and alleviating the vanishing gradient issue through the experiments in Section \ref{experiment}. 
  The results show that with deepening, our model achieves a significant improvement. Moreover, we take one step further to provide a reference of its deepening effect by comparing it with the state-of-the-art deepening technique ResNet proposed by He et al. \cite{He2015}. We take one ResNet block for illustration, which is shown in Fig.~\ref{ResDense}. The characteristic of ResNet is utilizing skip connections to add up the feature maps every two or three layers to encourage the information flow. A residual block has several different structures.  The one that we use here is ``bottleneck," which is one $3 \times 3$ convolution surrounded by dimensionality reducing and expanding $1 \times 1$ convolution layers \cite{ZagoruykoS}. For the test configuration, we simply change the DenseNet blocks in the multi-scale streams to the ResNet ``bottleneck" blocks and use max pooling operation instead of average pooling. We provide a comprehensive analysis of the deepening effect performance of both methods in the Section \ref{experiment}.

\subsection{Recovering the depth map}

Reconstructing the depth map from the estimates is to infer the global relationship of the selected points (we use each selected point to represent its superpixel) from the local relationship of point pairs. It is challenging since a considerable part of the estimated relationship between the points can be contradictory or ambiguous. We adopt the method proposed by Zoran et al.\ \cite{Zoran_2015_ICCV} to find the global solution, which poses this as a constrained quadratic optimization problem. In contrast to their approach, we solve the problem in log-space and introduce a smoothness term which is a reasonable prior for the task. Our objective is as follows:
\begin{equation}
\begin{aligned}
& \underset{\boldsymbol{x},\boldsymbol{\epsilon}}{\text{minimize}}
& & \mathrm{E}(\boldsymbol{x},\boldsymbol{\epsilon}) \\
& \text{subject to}
& & \boldsymbol{x} > L, \boldsymbol{x} < U, \boldsymbol{\epsilon} > 0
\end{aligned}
\label{eq1}
\end{equation}
where $E(\boldsymbol{x},\boldsymbol{\epsilon})$ is the energy function as follows,
\begin{equation}
E(\boldsymbol{x},\boldsymbol{\epsilon}) = \sum_{ij}\sum_{o}\omega_{o,i,j}\theta_{o}(x_i,x_j,\epsilon) + L_{s}(\boldsymbol{x}) + R(\boldsymbol{\epsilon})
\end{equation}
where $o\in \{=,>,<\}$, $\boldsymbol{x}$ are the depth values of the selected pixels. 
$\omega_{o,i,j}$ is the depth ordinal estimation (the outputs of the proposed model which are the probabilities for three cases) of the $ij$-th pair. $\epsilon \sim \mathcal{N}(\boldsymbol{\mu},\boldsymbol{\sigma^2})$ is a scalar slack variable for the ij-th pair.
 $\theta_{o}(x_i,x_j,\epsilon)$ is $L_2$ distance which penalizes the depth estimate when it disagrees with the estimates of the proposed model, which consists of
\begin{equation}
\begin{split}
\theta_{=}(x_i,x_j,\epsilon)= (|\log x_i - \log x_j|-\epsilon_{=,i,j})^2\\
\theta_{>}(x_i,x_j,\epsilon)= (\log x_i - \log x_j -\epsilon_{>,i,j})^2\\
\theta_{<}(x_i,x_j,\epsilon) = (\log x_j - \log x_i - \epsilon_{<,i,j})^2.
\end{split}
\end{equation}
$L_{s}(\boldsymbol{x})$ is the proposed smoothness term for the adjacent superpixels:
\begin{equation}
L_{s}(\boldsymbol{x}) = \sum_{ij}\omega_{i,j}(x_i-x_j)^2.
\end{equation}
It is weighted by the sum of local image gradient and the estimate of  `equal' case of the adjacent superpixels, which is
\begin{equation}
\omega_{i,j} = k_{1} \exp(-\frac{1}{\rho}||\boldsymbol{I_i} - \boldsymbol{I_j}||^2) + k_{2} \omega_{=}(x_i,x_j,\epsilon)
\end{equation}
where $\rho$ controls the sensitivity of the image gradient-based weight, $k_1$ and $k_2$ control the proportion of the two terms in the above equation.

 In the experiment, we set $k_1=k_2=0.5$. Note that, we only consider adjacent superpixels. For non adjacent neighors,
$\omega_{i,j}=0$.
The last term $R(\boldsymbol{\epsilon})$ is a regularization term to bound the $\boldsymbol{\epsilon}$, which is
\begin{equation}
R(\boldsymbol{\epsilon}) = \sum_{ij}(\frac{\epsilon_{=,i,j} - \mu_{=}}{\sigma_{=}^2} + \frac{\epsilon_{>,i,j} - \mu_{>}}{\sigma_{>}^2} + \frac{\epsilon_{<,i,j} - \mu_{<}}{\sigma_{<}^2}),
\end{equation}
where the mean values $\mu_{=}, \mu_{>}, \mu_{<}$ and the variances $\sigma_{=}^2, \sigma_{>}^2, \sigma_{<}^2$ are computed from the statistics of the training set,
 corresponding to `equal', `further' and `closer' cases, respectively.
 
  For the objective Equ.~\eqref{eq1}, $L, U$ are the lower and upper bonds for the depth values. Once the objective is solved, we generate the depth map by floodfilling each superpixel with the corresponding values.

\begin{figure*}[t]
\centering
\begin{minipage}{0.19\textwidth}
{\includegraphics[width=\linewidth]{2.png}}
\end{minipage}
\begin{minipage}{0.19\textwidth}
{\includegraphics[width=\linewidth]{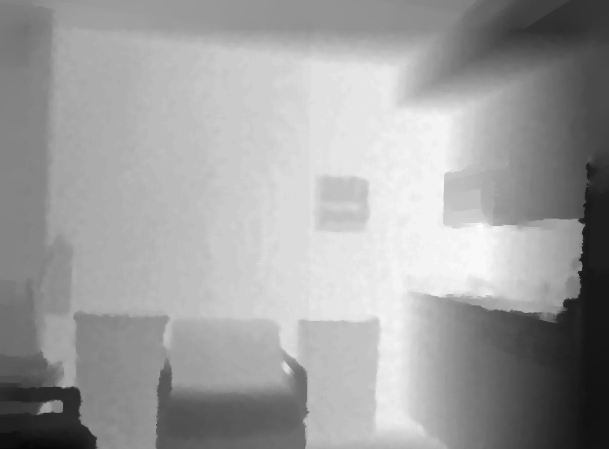}}
\end{minipage}
\begin{minipage}{0.19\textwidth}
{\includegraphics[width=\linewidth]{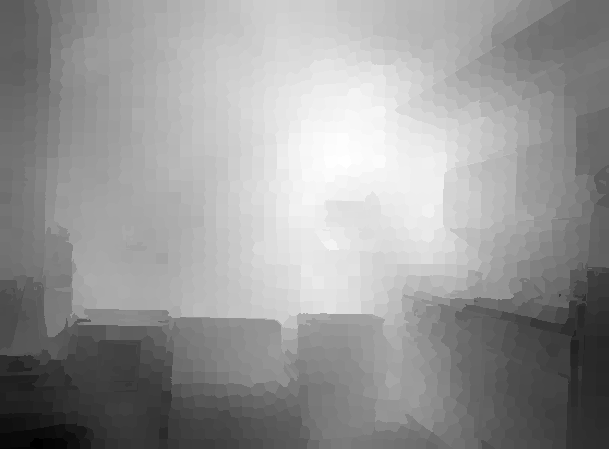}}
\end{minipage}
\begin{minipage}{0.19\textwidth}
{\includegraphics[width=\linewidth]{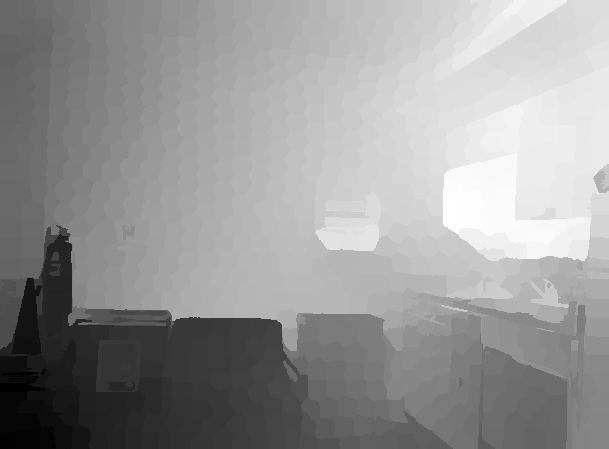}}
\end{minipage}
\begin{minipage}{0.19\textwidth}
{\includegraphics[width=\linewidth]{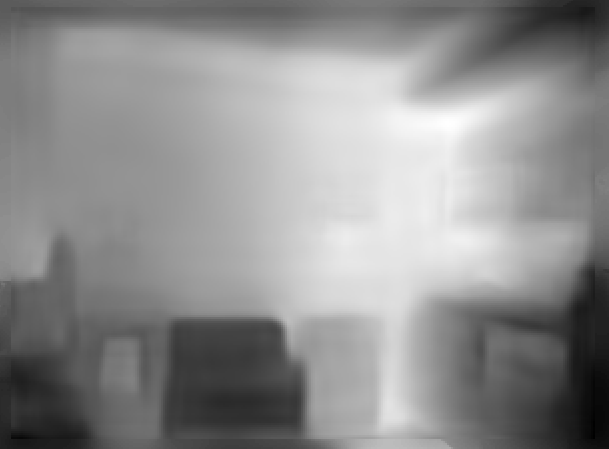}}
\end{minipage}
\vskip 0.05in
\begin{minipage}{0.19\textwidth}
{\includegraphics[width=\linewidth]{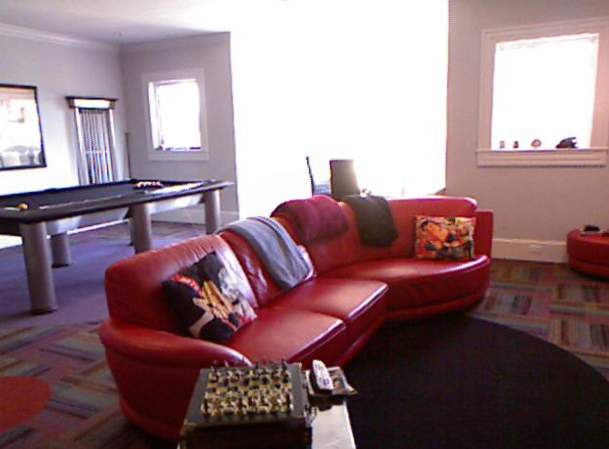}}
\end{minipage}
\begin{minipage}{0.19\textwidth}
{\includegraphics[width=\linewidth]{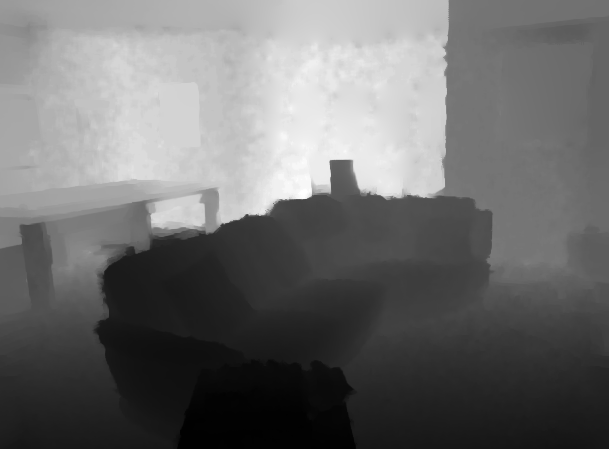}}
\end{minipage}
\begin{minipage}{0.19\textwidth}
{\includegraphics[width=\linewidth]{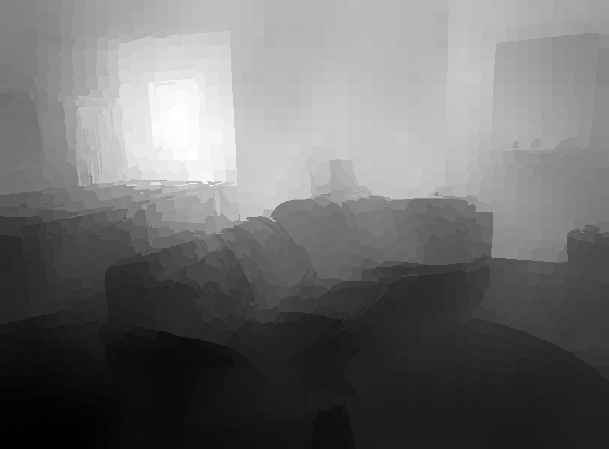}}
\end{minipage}
\begin{minipage}{0.19\textwidth}
{\includegraphics[width=\linewidth]{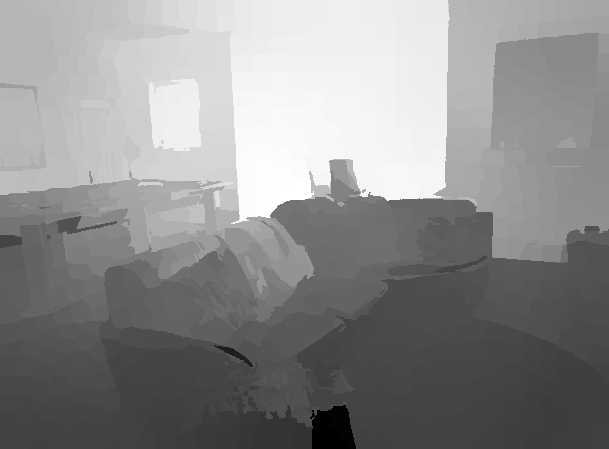}}
\end{minipage}
\begin{minipage}{0.19\textwidth}
{\includegraphics[width=\linewidth]{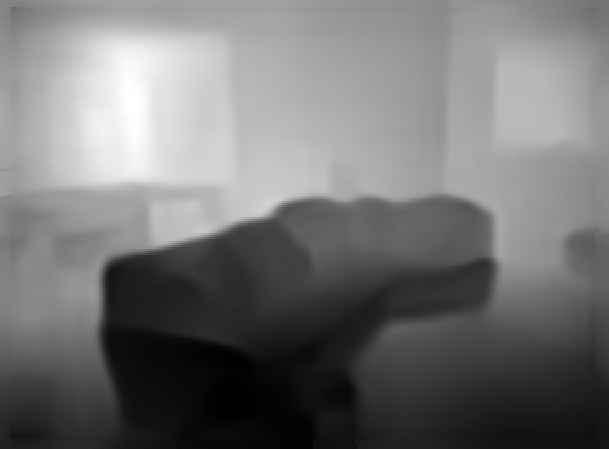}}
\end{minipage}
\vskip 0.05in
\begin{minipage}{0.19\textwidth}
{\includegraphics[width=\linewidth]{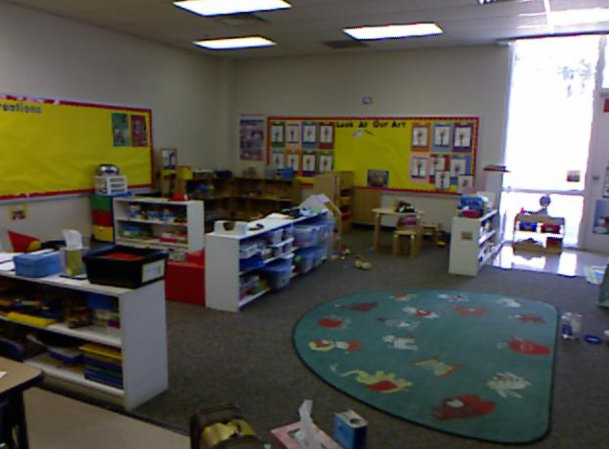}}
\end{minipage}
\begin{minipage}{0.19\textwidth}
{\includegraphics[width=\linewidth]{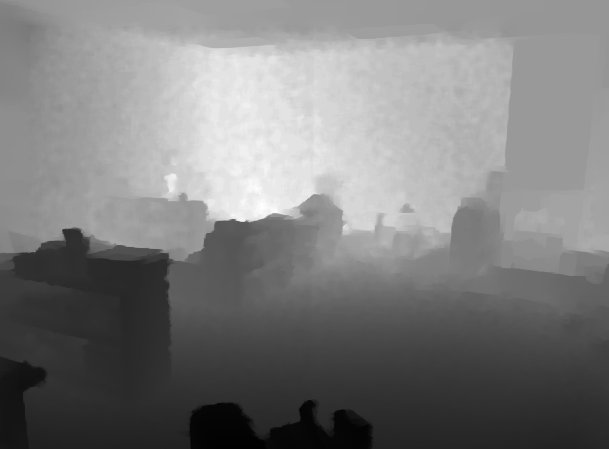}}
\end{minipage}
\begin{minipage}{0.19\textwidth}
{\includegraphics[width=\linewidth]{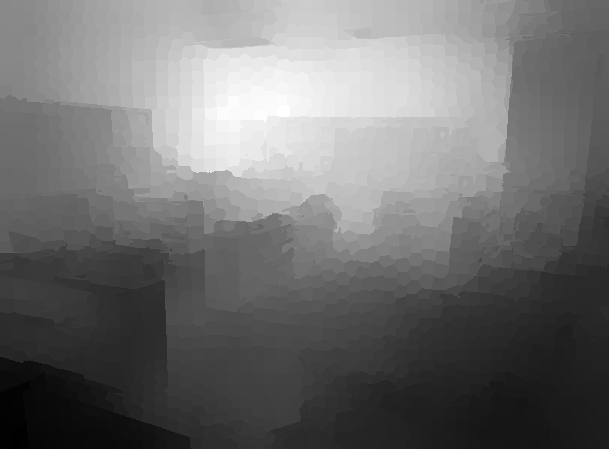}}
\end{minipage}
\begin{minipage}{0.19\textwidth}
{\includegraphics[width=\linewidth]{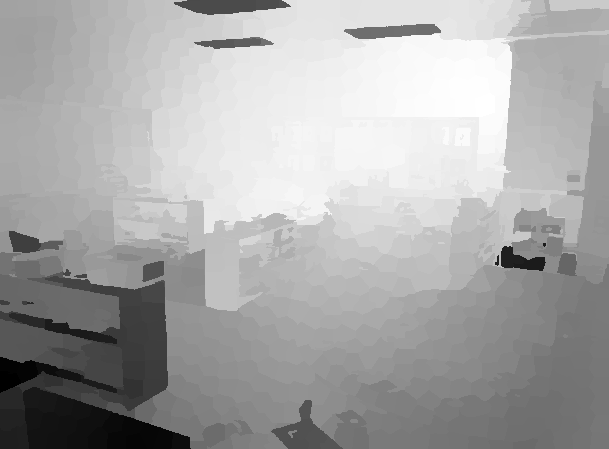}}
\end{minipage}
\begin{minipage}{0.19\textwidth}
{\includegraphics[width=\linewidth]{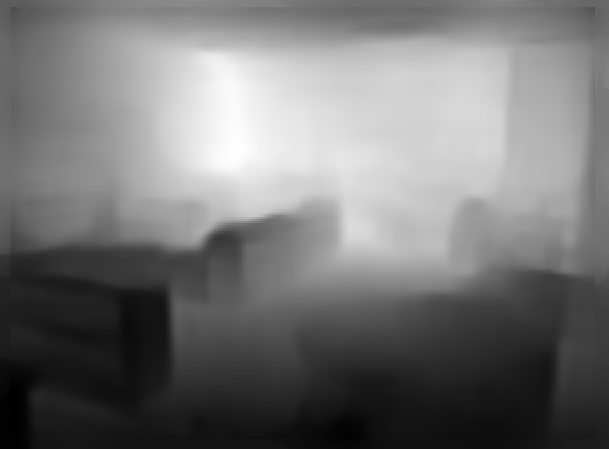}}
\end{minipage}
\vskip 0.05in
\begin{minipage}{0.19\textwidth}
\subfloat[Original Image]{\label{main:a}\includegraphics[width=\linewidth]{4.png}}
\end{minipage}
\begin{minipage}{0.19\textwidth}
\subfloat[GroundTruth]{\label{main:e}\includegraphics[width=\linewidth]{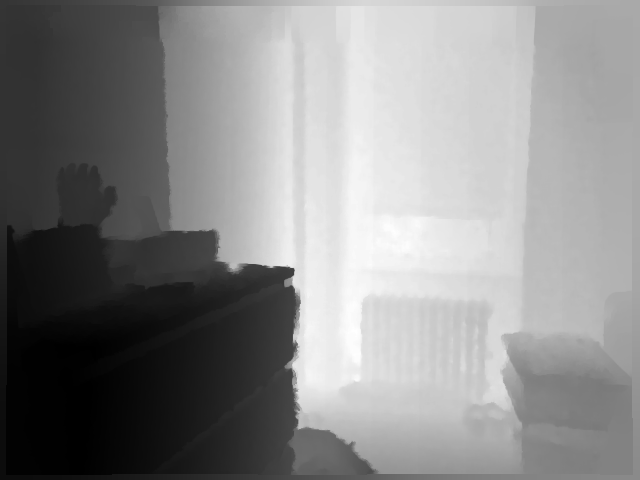}}
\end{minipage}
\begin{minipage}{0.19\textwidth}
\subfloat[Ours]{\label{main:b}\includegraphics[width=\linewidth]{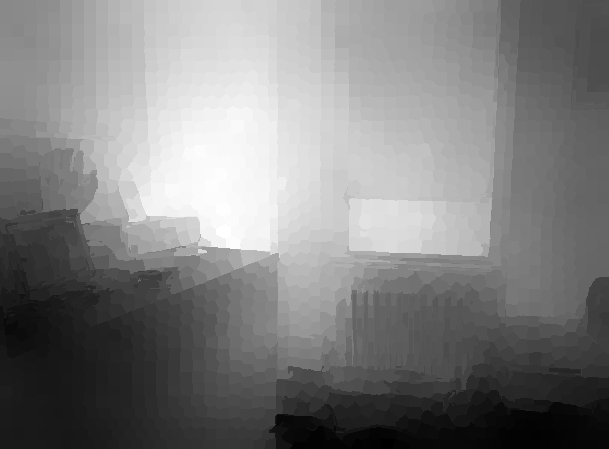}}
\end{minipage}
\begin{minipage}{0.19\textwidth}
\subfloat[Zoran et al. \cite{Zoran_2015_ICCV}]{\label{main:c}\includegraphics[width=\linewidth]{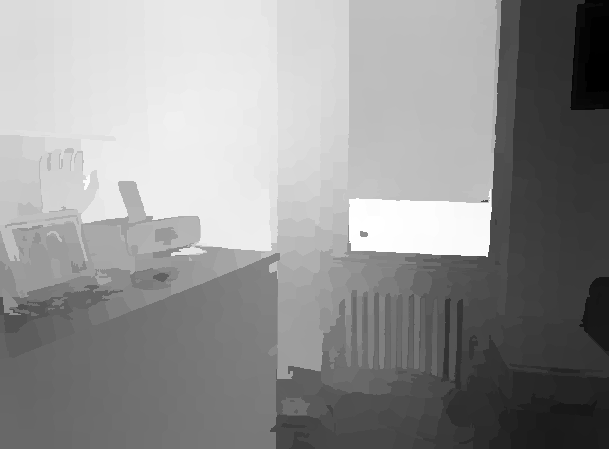}}
\end{minipage}
\begin{minipage}{0.19\textwidth}
\subfloat[Eigen et al. \cite{Eigen_2015_ICCV}]{\label{main:d}\includegraphics[width=\linewidth]{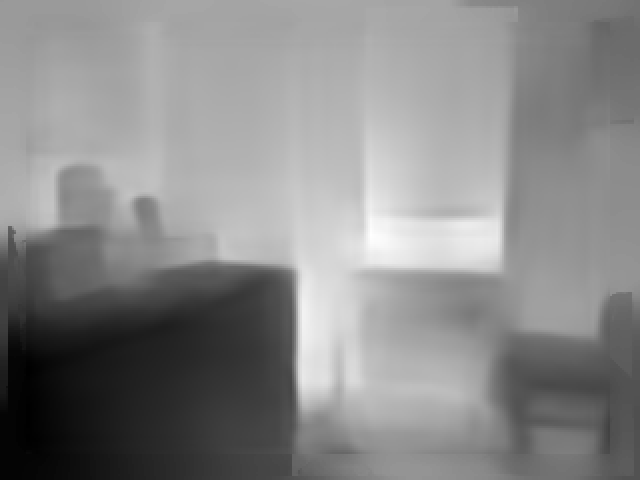}}
\end{minipage}

\caption{Recovered depth maps from single images on NYU depth dataset. The darker intensity indicates the pixel is closer to the camera. The lighter intensity indicates the pixel is further to the camera.}
\label{fig:main2}
\end{figure*}

\section{Experiment}
\label{experiment}
We demonstrate the effectiveness of the proposed model on three datasets including the 
NYU depth V2 dataset \cite{silberman2012indoor}, KITTI dataset \cite{geiger2013vision} and DIW dataset \cite{ChenFYD16}. 
The experiments are organized in two categories:
 
 1) We compare the proposed model with several baseline methods to show the benefits of integrating the multi-scale features and DenseNet structure; 
 
 2) We compare the proposed model with the state-of-the-art methods on each dataset to show that our method performs better.

\textbf{Training settings} The implementation is in Caffe \cite{Jia2014}. The contexts are extracted and resized to the specific resolutions 
as the inputs to the network. We show the size of each input in Table \ref{training setting 2}. The kernel size, the stride, the outputs number of each layer are reported in Table \ref{training setting 2}. Note that for the streams of the scales 1, 2, 3 in Fig.~\ref{network structure}, the `layer 1/2/3/4' means each convolutional layer before the corresponding DenseNet blocks. For each DenseNet block, $3\times3$ kernel size with zero padding are used in it to keep the feature size fixed and the growth $k$ of the DenseNet is 5.
\begin{table}[t!]
\renewcommand{\arraystretch}{1.23}
\caption{ The size of input contexts.}
\centering
\resizebox{\columnwidth}{!}{
\begin{tabular}{ c||c|c|c|c|c }
 \hline
Input Context & P1/P2& Scale 1 & Scale 2 & Scale 3& Mask\\
 \hline
 Size & $16*16$ & $32*32$ & $40*40$ & $48*48$ & $32*32$\\
\hline
\end{tabular}
}
\label{training setting 1}
\end{table}

\begin{table}[b!]
\renewcommand{\arraystretch}{1.23}
\newcommand{\specialcell}[2][c]{%
  \begin{tabular}[#1]{@{}c@{}}#2\end{tabular}}
\caption{ The kernel size, stride, outputs number and pooling size of each layer in each stream.}
\centering
\resizebox{\columnwidth}{!}{
\begin{tabular}{ c||c|c }
 \hline
Stream & P1/P2& Scale 1/Scale 2/Scale 3\\
\hline
 layer 1 & $3*3,2,24$,no pooling & \specialcell{ $3*3,1$(zero padding),$24$,\\$2*2$ average pooling}\\
\hline
 layer 2 & $3*3,1,24$,no pooling & \specialcell{ $3*3,1$(zero padding),$24$,\\$2*2$ average pooling}\\
\hline
 layer 3 & $3*3,2,40$,no pooling & \specialcell{ $3*3,1$(zero padding),$48$,\\$2*2$ average pooling}\\
\hline
 layer 4 & $3*3,1,40$,no pooling &  \specialcell{ $3*3,1$(zero padding),$48$,\\$2*2$ average pooling}\\
\hline
\end{tabular}
}
\label{training setting 2}
\end{table}

The network is learned end-to-end using a log-softmax loss. We train the model from scratch and use stochastic gradient descent (SGD)  for optimization. 
For NYU V2 and KITTI depth datasets, we trained 300k iterations with mini-batches of 256 pairs, for DIW, 400k iterations with mini-batches of 128 pairs. The weight decay is 0.0005. We use an NVIDIA GeForce Titan X GPU, on which training process takes roughly 5 hours. For the ablation experiments, we reduce the training iterations to 200k. The other settings keep the same.

\textbf{Metric error measure}
 We utilize  Weighted Kinect Disagreement Rate (WKDR) \cite{Zoran_2015_ICCV} metric, which is an average disagreement rate, to evaluate all the methods' performance.

\[
\mathbf{WKDR}_\delta(l_i,\boldsymbol{x}) = \frac {\sum_{ij}\mathbbm{1} (l_i \neq l_{i,  \delta}(\boldsymbol{x}))} {\sum_{ij}}
\]

\[
l_{ij, \delta}(\boldsymbol{x}) = \begin{cases}
      1 &  \text{if} \:  \frac{x_i} {x_j} > 1+ \delta\\
      2 &  \text{if}\:  \frac{x_j} {x_i} > 1+ \delta\\
      E &  \text{else}
   \end{cases}
\]
where $\textbf{x}$ is the estimated depth map. 
We set the tolerance level $\delta = 0.02$ which is the same with \cite{Zoran_2015_ICCV}.
 $l_{ij}$ is the annotation of the depth order of $ij$-th pair. $x_i,x_j$ are the depth values of two points in a pair.

\textbf{Baseline settings} To demonstrate the effectiveness of the multi-scale features and the deepening technique, we present three baseline methods:
\begin{itemize}
\item Baseline A: The first baseline is a single scale model without deepening. It is a simplified version of the proposed basic model in the Section \ref{proposed}, which removes the scale 2, scale 3 streams. The method is expected to show the fundamental performance without multi-scale architecture and any deepening method.
\item Baseline B: The second baseline is the single scale model with deepening (DenseNet block added in the scale 1 stream). By comparing with Baseline A, we will see the difference of the single scale model before and after deepened.
\item Baseline C: The last baseline is the proposed basic model in Section \ref{proposed}, i.e. multi-scale without deepening. By comparing with Baseline B, we can explore as to whether deepening or using a multi-scale architecture provides the bigger improvement in performance.
\end{itemize}
We carry out the baseline comparisons compared against the proposed model  on all the datasets.

\subsection{NYU Depth Dataset}
The NYU depth dataset \cite{silberman2012indoor} is a large depth benchmark for indoor scenes, which is collected by a Microsoft Kinect sensor.
 It consists of 464 indoor scenes and more than 400k images with the resolution of $480\times640$. 
 We use its densely labeled dataset, which has 1449 pairs of aligned RGB and depth images. 
 We use its official train/test split, that is 795 images for training, 654 images for testing. We sample 1600 pairs per image for training and 800 pairs per image for testing. That is totally 127.2 million pairs of points for training and 52.3 million pairs for testing. We consider the number of the training data enough thus we do not apply any data augmentation.

\textbf{Baseline comparisons} The results are reported in Table \ref{baseline comparisons nyu}. 
As we can see, applying the multi-scale features without deepening provides larger performance improvement  than deepening on a single stream. Meanwhile deepening the streams of the model is very useful to boost the performance. Among the methods, our model performs the best.

\begin{table}[h]
\renewcommand{\arraystretch}{1.23}
\caption{The baseline comparisons on the NYU depth dataset. The proposed deepened model performs the best.}
\centering
\begin{tabular}{ c||c|c|c  }
 \hline
 Method & WKDR &$\text{WKDR}^{=}$&$\text{WKDR}^{\ne}$\\
 \hline
 \text{Baseline A}  & 38.0\%  &36.7\%   &39.4\%\\
 \text{Baseline B}  & 36.8\%  &35.2\%   &38.9\%\\
 \text{Baseline C}  &35.9\% & 34.5\%&  37.1\%\\
 \text{The proposed model}  & \textbf{33.9}\% & \textbf{32.1}\% & \textbf{38.5}\% \\
\hline
\end{tabular}
\label{baseline comparisons nyu}
\end{table}

\begin{figure*}[t]
\centering
\begin{minipage}{0.325\textwidth}
{\includegraphics[width=\linewidth]{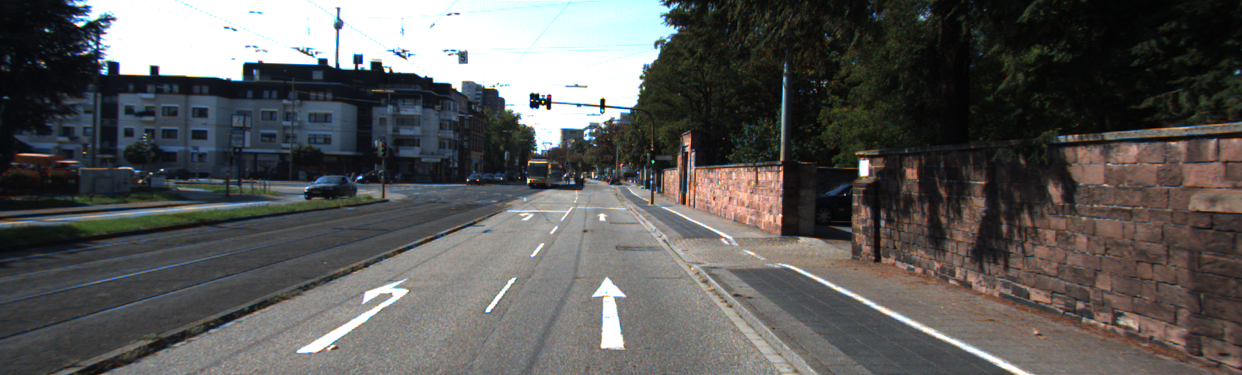}}
\end{minipage}
\begin{minipage}{0.325\textwidth}
{\includegraphics[width=\linewidth]{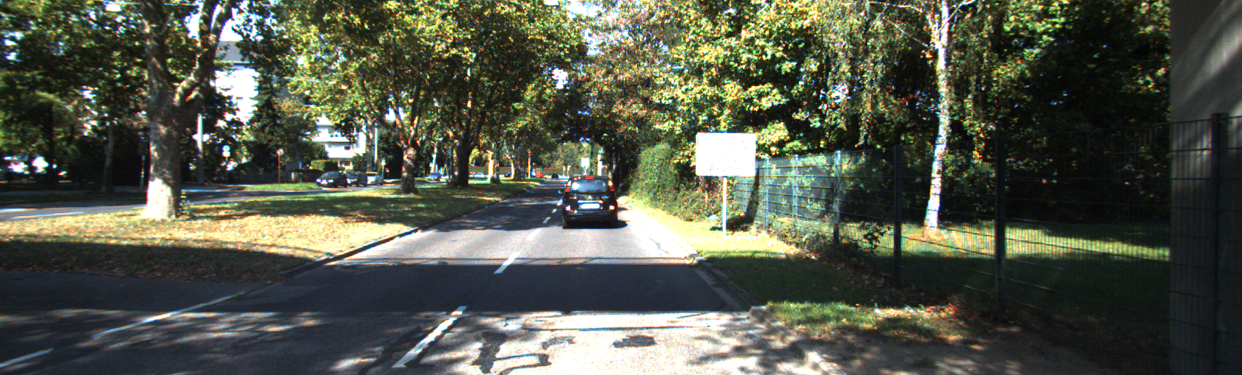}}
\end{minipage}
\begin{minipage}{0.325\textwidth}
{\includegraphics[width=\linewidth]{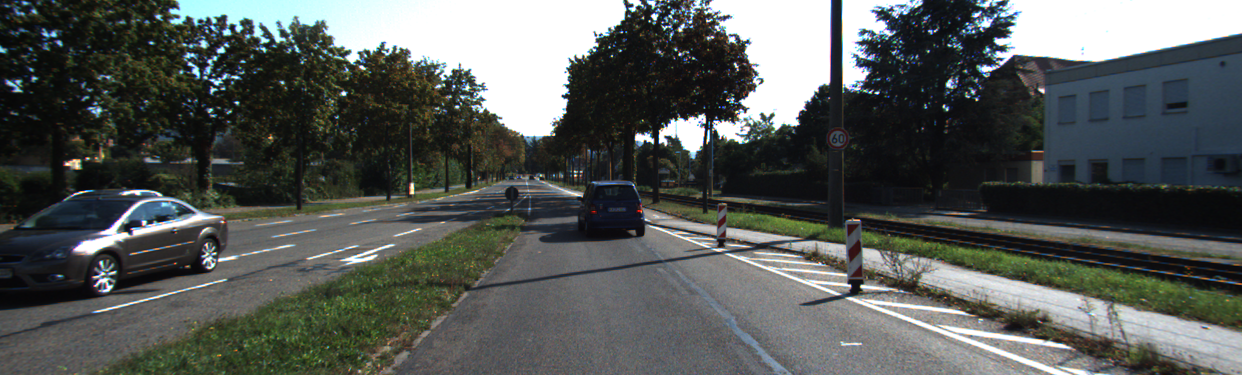}}
\end{minipage}

\vskip 0.05in
\begin{minipage}{0.325\textwidth}
{\includegraphics[width=\linewidth]{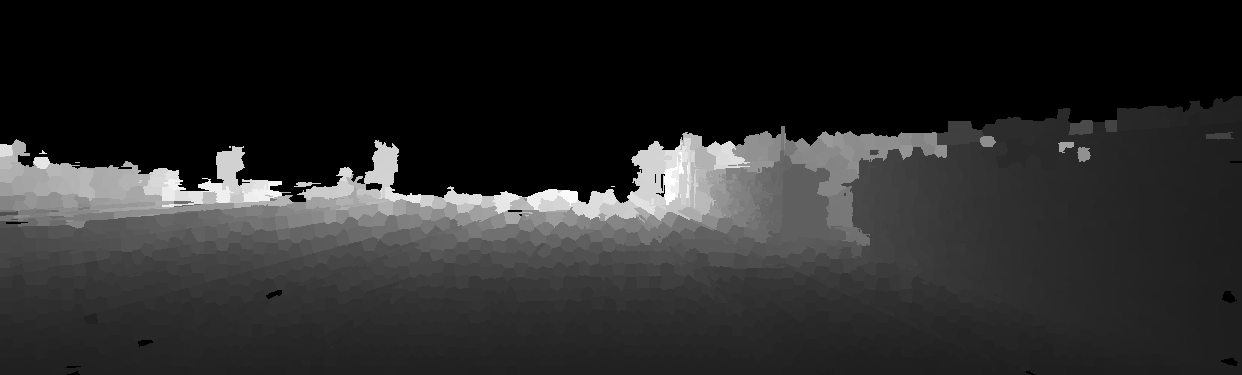}}
\end{minipage}
\begin{minipage}{0.325\textwidth}
{\includegraphics[width=\linewidth]{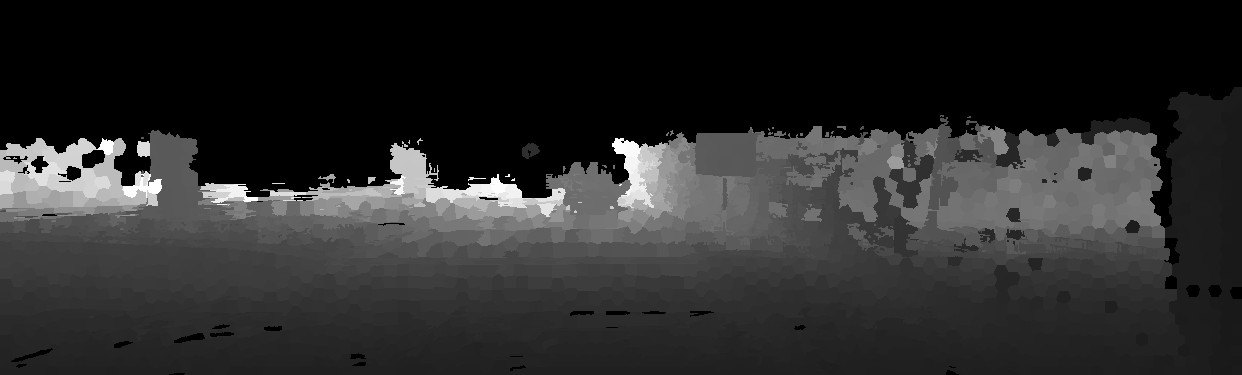}}
\end{minipage}
\begin{minipage}{0.325\textwidth}
{\includegraphics[width=\linewidth]{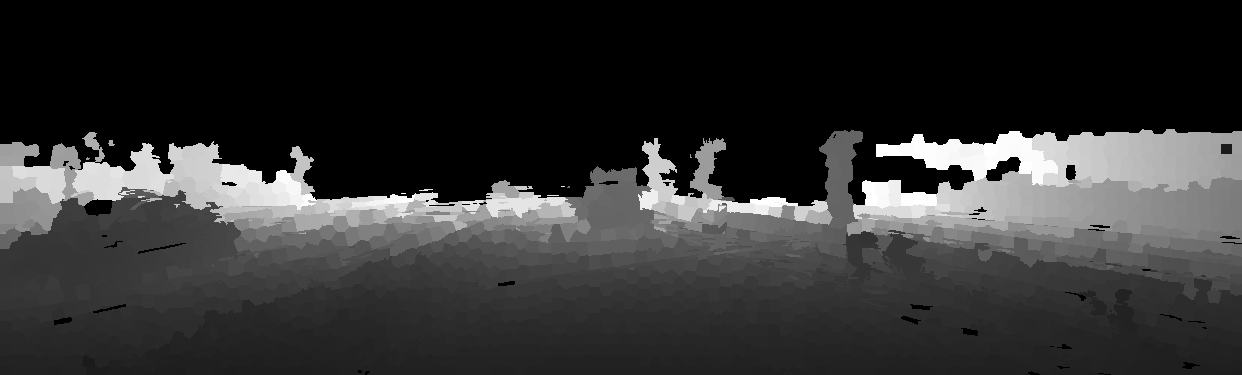}}
\end{minipage}

\vskip 0.05in
\begin{minipage}{0.325\textwidth}
{\includegraphics[width=\linewidth]{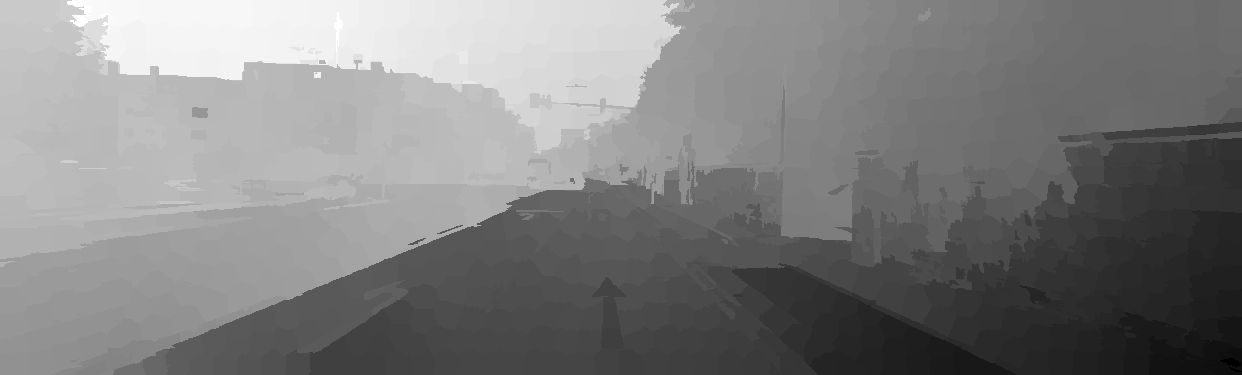}}
\end{minipage}
\begin{minipage}{0.325\textwidth}
{\includegraphics[width=\linewidth]{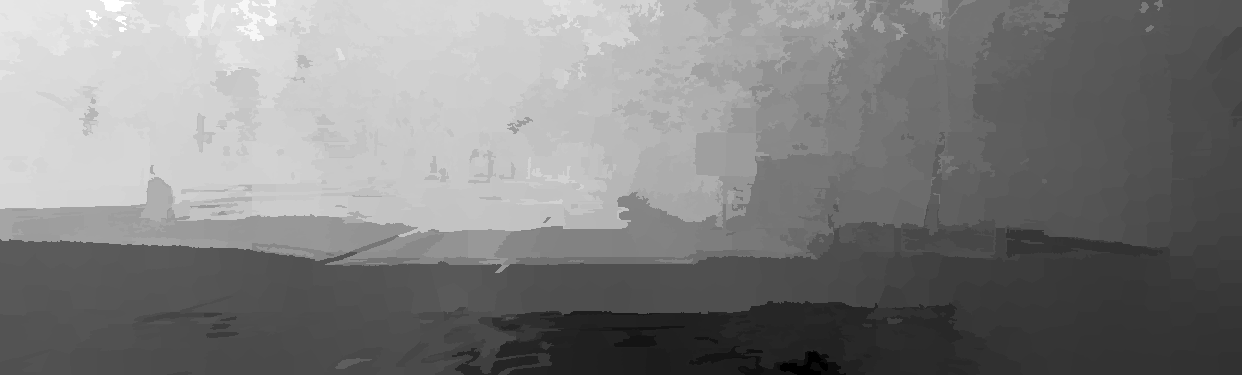}}
\end{minipage}
\begin{minipage}{0.325\textwidth}
{\includegraphics[width=\linewidth]{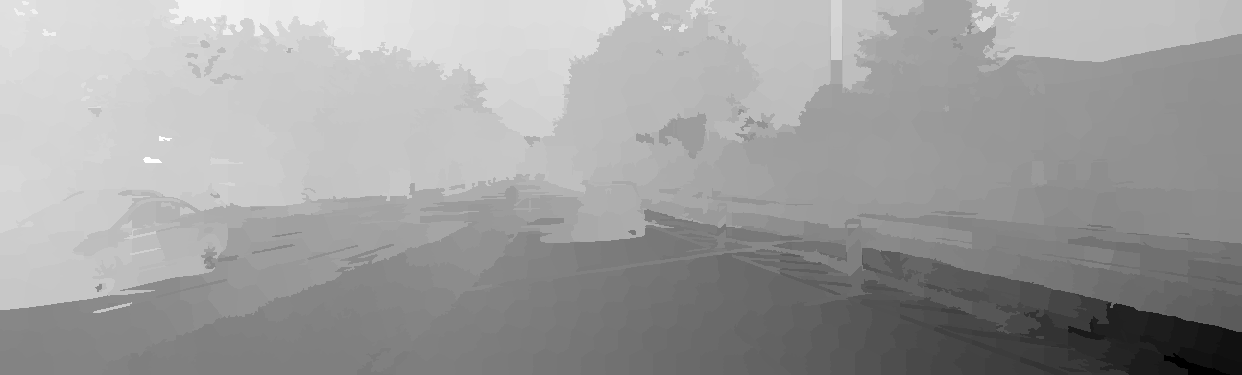}}
\end{minipage}

\vskip 0.05in
\begin{minipage}{0.325\textwidth}
\subfloat{\includegraphics[width=\linewidth]{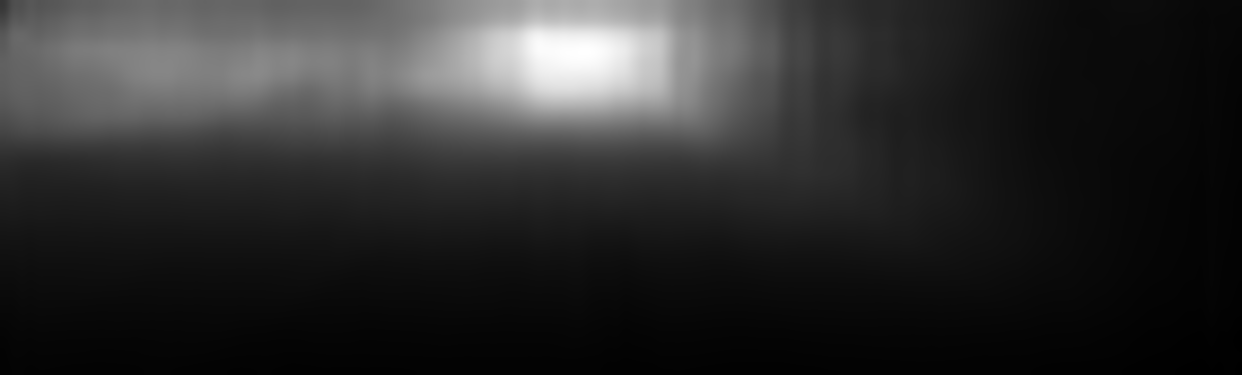}}
\end{minipage}
\begin{minipage}{0.325\textwidth}
\subfloat{\includegraphics[width=\linewidth]{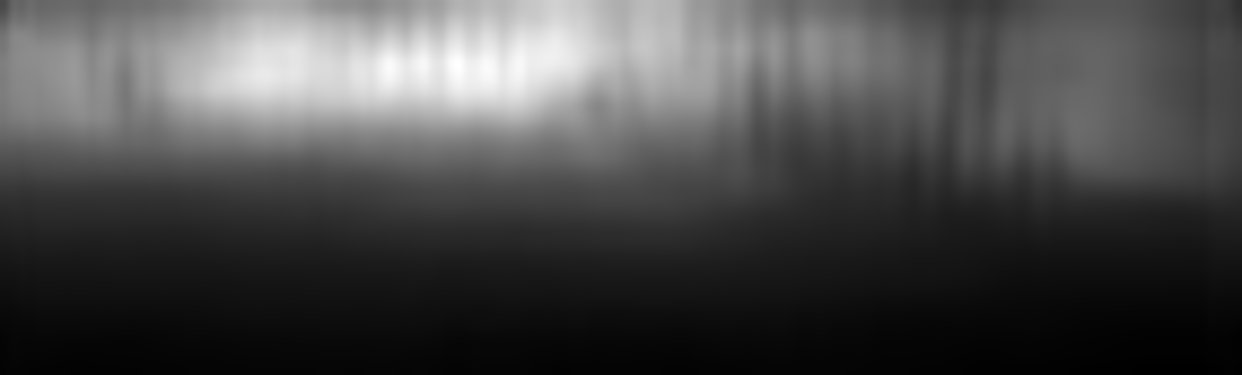}}
\end{minipage}
\begin{minipage}{0.325\textwidth}
\subfloat{\includegraphics[width=\linewidth]{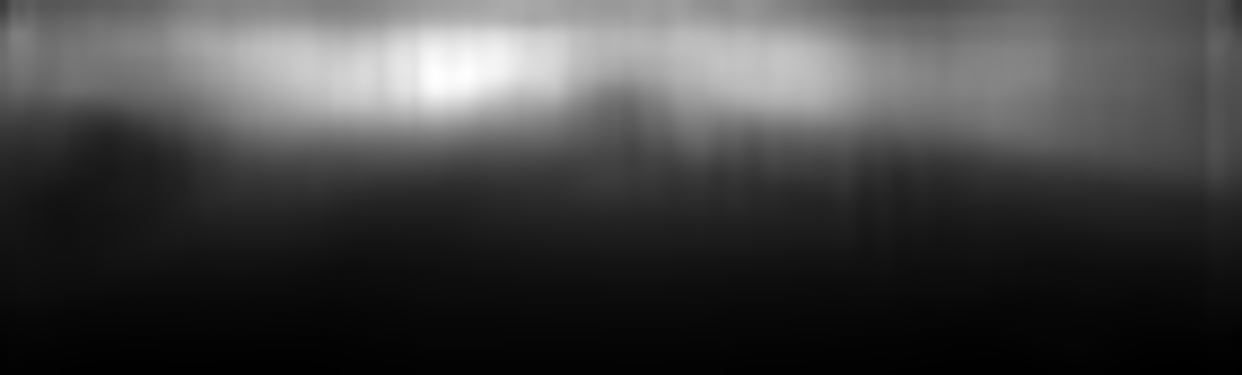}}
\end{minipage}

\caption{Three examples on KITTI dataset. From the top row to the bottom row is the original images, groundtruth, our results, Eigen et al.'s results.}
\label{kitti}
\end{figure*}

\textbf{Comparisons of deepening methods } The second experiment is the comparison test of applying two deepening technique according to Section \ref{deepened}. The basic model is the proposed multi-scale model without deepening, which we used as Baseline C. As shown in Table \ref{res and dense}, the methods both help improve the performance. Yet DenseNet outperforms than ResNet. In the training process of  ResNet, we observe the overfitting phenomenon. We hypothesize the issue is caused by the duplicated features and the excessive filter numbers of the ResNet block. The three scale features are duplicated due to the overlapping area of the contexts (they overlap at the scale 1 area). In the case, the model needs the regularization to avoid the risk of overfitting the duplicated feature space. However, the `bottleneck' structure of the ResNet has 64, 64, 256 filters in a block, whose total filter number is six times more than a DenseNet block that we use, which means that the ResNet has a greater chance to overfit the data than the DenseNet. Thus we use the DenseNet model, and the experimental result demonstrates that 
the method works well for the proposed model.

\begin{table}[h]
\caption{Deepening methods comparison on the NYU depth dataset.}
\centering
\begin{tabular}{ c||c  }
 \hline
 Method &  WKDR\\
\hline
 The basic model & 35.9\% \\

 ResNet & 35.1\%\\

 DenseNet & \textbf{33.9}\% \\
 \hline
\end{tabular}
\label{res and dense}
\end{table}

\textbf{State-of-the-art comparisons}  The state-of-the-art methods for comparison on the dataset are Eigen et al.\ \cite{Eigen_2015_ICCV}, Zoran et al.\ \cite{Zoran_2015_ICCV} and Chen et al.\ \cite{ChenFYD16}. Note that, all the methods train and test their models on the densely labeled data, yet Chen and Eigen additionally use the raw dataset which contains more than 290k sparsely labeled depth images to achieve the better performance. 

Since the numbers of training images in the densely labeled data and the raw NYU dataset are vastly different, the performances of the models trained on each dataset are reported separately in Table \ref{nyu}. We first focus on the upper part of the table which shows the results of the models trained on the densely labeled data. An interesting observation from the table is that our one scale model (baseline B) already significantly outperforms than Zoran et al.'s result and is slightly better than Chen et al. \cite{ChenFYD16}. Note that Chen et al.\ apply more than 12k points pairs per image for training, while we only need 1.6k pairs. Our multi-scale model achieves the state-of-the-art performance on the densely labeled dataset.

\begin{table}[b!]
\renewcommand{\arraystretch}{1.23}
\newcommand{\specialcell}[2][c]{%
  \begin{tabular}[#1]{@{}c@{}}#2\end{tabular}}
\caption{State-of-the-art comparisons on the NYU depth dataset.}
\centering
\begin{tabular}{ c||c|c|c  }
\hline
\multicolumn{4}{c}{\specialcell{ Trained on the NYU densely labeled data\\(795/654, train/test split)} } \\
 \hline
 Method & WKDR &$\text{WKDR}^{=}$&$\text{WKDR}^{\ne}$\\
 \hline
 \text{Zoran et al. \cite{Zoran_2015_ICCV}} &   43.5\%  &44.2\%   &41.4\%\\
 \text{Chen et al.\ \cite{ChenFYD16}}   &38.7\% & 39.7\%&  39.4\%\\
\text{Ours}  & \textbf{33.9}\%    &\textbf{32.1}\%&   \textbf{35.4}\%\\
\hline
\multicolumn{4}{c}{\specialcell{ Trained on the raw NYU dataset\\(290k/654, train/test split)}} \\
 \hline
 Method & WKDR &$\text{WKDR}^{=}$&$\text{WKDR}^{\ne}$\\
\hline
 \text{Chen et al.\ \cite{ChenFYD16}}   &\textbf{28.7\%} & \textbf{31.2\%}&  \textbf{28.7\%}\\
 \text{Eigen(V)\cite{Eigen_2015_ICCV}} &34.0\% &43.3\%&  29.6\%\\
 \hline
\end{tabular}
\label{nyu}
\end{table}

The bottom part of the table shows the results of the models trained on the raw NYU dataset. Although the number of training data 
is very different,  the performance of the proposed model is still comparable to Eigen et al.\ \cite{Eigen_2015_ICCV}. The improvement of Chen et al.'s model from two datasets demonstrates the large training data can greatly boost the performance of the DCNN.

 However, for the proposed model, training on the raw NYU dataset can be very time-consuming and also needs massive hard disk space. Because the time of generating the contexts of the points pairs and converting these small image patches to the data format that CAFFE can fast read increases with the training data. We believe the results of training on the densely labeled data have already demonstrated the superiority of the proposed model.

In Fig.~\ref{fig:main2}, we show four example depth maps reconstructed from the depth ordinal predictions. Compared with Zoran et al.\ \cite{Zoran_2015_ICCV}, our results (Column 3)  demonstrate more accurate reconstruction. For instance, in the first example, the furthest region reconstructed by Zoran et al.\ is the cabinet on the upper right corner. We manage to find the correct area which is the wall on the left side of the cabinet. 
The other advantage of our results is the smoothness, such as the three ceiling lamps in the third example, compared with Zoran et al.

\subsection{KITTI dataset}
The KITTI dataset \cite{geiger2013vision} is a large and comprehensive dataset for benchmarking the autonomous driving techniques. It contains a number of outdoor scenes for depth estimation, which has five broad categories: `City,' `Residential,' `Road,' `Campus' and `Person'.

 We choose `City' raw dataset for evaluation which includes 27 scenes. We use the train-test split index of the scenes provided by Eigen \cite{Eigen_2015_ICCV} in the `City' category, i.e., 18 scenes for training and 9 scenes for testing. 
 
 The dataset has no direct annotations for the task. Therefore we use the same method as we did in the NYU depth dataset to generate the annotations. Since the data of each scene is a video sequence, the content of each frame in the same scene has a lot of redundancy.
  Thus we sparsely sample 200 points pairs per image, which is much less than the number in the NYU dataset to avoid the duplicate pairs.
   We gather in total 815k pairs for training and 450k pairs for testing. Note that, the ground-truth depths of the KITTI dataset are scattered at irregularly spaced points, which only consists of $\sim 5\%$ pixels of each image, we extract the ground-truth depth closest to each superpixel centroid as the superpixel depth and floodfill the superpixels with the relative depth values.

\textbf{Baseline comparisons}
The baseline comparisons on the KITTI dataset are reported in Table \ref{baseline comparisons kitti}. We observe two interesting phenomena. 

Firstly, the performances of predicting the equal case and the unequal case are very different. We argue that it is caused by learning the data of unbalanced distribution. The structure of the road scene is that a road is always in the center and the buildings on the two sides are along the road, which makes the KITTI dataset has a distinct pattern that the ground truth depth increases with the road's direction. Owing to the pattern, the unequal case accounts for the most of the points pairs, roughly $80\% \sim 90\%$ of the total number of each image. Thus the unequal case is learned very well, yet the equal case is not. 

Secondly, with the multi-scale feature and the DenseNet block added, the performance of predicting the equal case show a much  more significant improvement than the unequal case.
\begin{table}[b!]
\renewcommand{\arraystretch}{1.23}
\caption{The baseline comparisons on the KITTI dataset.}
\centering
\begin{tabular}{ c||c|c|c  }
 \hline
 Method & WKDR &$\text{WKDR}^{=}$&$\text{WKDR}^{\ne}$\\
 \hline
 \text{Baseline A}  & 30.4\%  &79.9\%   &23.3\%\\
 \text{Baseline B}  & 29.3\%  &74.1\%   &22.9\%\\
 \text{Baseline C}  &27.3\% &71.9\%&  21.4\%\\
 \text{The proposed model}  & \textbf{25.8}\% & \textbf{66.9}\% & \textbf{20.6}\% \\
\hline
\end{tabular}
\label{baseline comparisons kitti}
\end{table}

\textbf{State-of-the-art comparisons} We compare the performance with the state-of-the-art method Eigen et al. \cite{Eigen_2015_ICCV}.  The proposed method achieves the state-of-the-art performance and the recovered examples in Fig.~\ref{kitti} show that the overall structures and the crisp edges at depth discontinuities are captured.
\begin{table}[t!]
\caption{State-of-the-art comparisons on the KITTI dataset.}
\centering
\begin{tabular}{ c||c|c|c  }
 \hline
 Method & WKDR &$\text{WKDR}^{=}$&$\text{WKDR}^{\ne}$\\
\hline
 Eigen et al. & 26.3\% & \textbf{64.7}\%& 21.2\%\\
 Ours & \textbf{25.8}\% & 66.9\%& \textbf{20.6}\% \\
 \hline
\end{tabular}
\label{kitti wkdr}
\end{table}

\subsection{Depth In the Wild dataset}
The Depth in the Wild (DIW) dataset \cite{ChenFYD16} is a recently released dataset for relative depth estimation. The points pairs are human-annotated. Thus we do not need to manually pick the points and generate the labels. The resolutions of the images are not fixed, roughly at $500\times
400$. The dataset uses more than 421k images for training and 74k images for testing. For each image.
 It only annotates one pair of the points. Thus we have 421k pairs of the points for training and 74k pairs for testing.

Different from the previous datasets, the DIW dataset has two distinct characteristics:
 
 1) The selected points in each image has a relative longer distance between each other, compared with the points pairs generated in the NYU and KITTI datasets, and a lot of points pairs locates at the same row or column; 
 
 2) More importantly, the dataset only considers two cases, `closer' and `further', where the `equal' case is removed. Thus for the dataset, WKDR metric is equal to $\text{WKDR}^{\ne}$. For the first characteristic, we change the three-scale bounding boxes generating strategy: for the points pairs locating in the same row or column, we extract the three scale bounding boxes, whose height or width are 20 pixels, 40 pixels, and 60 pixels, centered on the line between the points; for the other points pairs, we use the same strategy reported in Fig.~\ref{boundingboxes}.

\textbf{Baseline comparisons}
The baseline comparisons on the DIW dataset are reported in Table \ref{baseline comparisons diw}. The proposed model outperforms the other baseline methods.
\begin{table}[b!]
\renewcommand{\arraystretch}{1.23}
\caption{The baseline comparisons on the KITTI dataset.}
\newcommand{\specialcell}[2][c]{%
  \begin{tabular}[#1]{@{}c@{}}#2\end{tabular}}
\centering
\begin{tabular}{ c||c|c|c|c  }
 \hline
 Method & \text{Baseline A} &\text{Baseline B} & \text{Baseline C}&\text{\specialcell{The proposed \\model}} \\
 \hline
 \text{WKDR}  & 26.1\%  &24.5\%   &23.6\%& \textbf{20.3}\% \\
\hline
\end{tabular}
\label{baseline comparisons diw}
\end{table}

\textbf{State-of-the-art comparisons}
We compare the proposed model with Chen et al. \cite{ChenFYD16}. We follow their method to show the performance in two ways:
 1) training the model from scratch; 
 2) pretrain on the NYU depth dataset then finetune on the DIW dataset. 
 
 The results are reported in Table \ref{diw}. The results show that each model has its own merits. Our model significantly outperforms Chen et al., when both are trained from scratch. While their model has made a considerable improvement when pretrain and finetune from the NYU depth raw dataset and achieved a much better result than ours. Their pretrain model is based on the NYU raw dataset which has a much lower error rate than our pretrained model (see Table \ref{nyu}) and gains powerful visual representation. The benefits help improve their result.

\begin{table}[t!]
\renewcommand{\arraystretch}{1.23}
\caption{WKDR of different algorithms on the DIW dataset.}
\newcommand{\specialcell}[2][c]{%
  \begin{tabular}[#1]{@{}c@{}}#2\end{tabular}}
\centering
\begin{tabular}{ c||c|c  }
 \hline
\multicolumn{3}{c}{ Training from scratch } \\
\hline
 Method & The proposed model & Chen et al. \cite{ChenFYD16}\\
 \hline
WKDR & \textbf{20.3}\%& 25.5\%\\
\hline
\multicolumn{3}{c}{ Finetuning from NYU } \\
\hline
 Method & The proposed model & Chen et al. \cite{ChenFYD16}\\
 \hline
WKDR & 19.7\%& \textbf{16.3}\%\\
 \hline
\end{tabular}
\label{diw}
\end{table}

\section{Conclusion}
Predicting the depth order of the points pairs is a challenging task. By effectively exploring the contexts surrounding the points and deepening the network, it can be performed very well. The proposed framework accomplishes this through the use of the multi-scale local contexts and the DenseNet technique. We achieve  state-of-the-art on the task for several datasets with the advantage of using much fewer training data. Future work will extend to solve  other mid-level vision issues such as intrinsic image decomposition and improve the framework to exploit more data.

\bibliographystyle{IEEEtran}
\bibliography{mybibfile}

\end{document}